# Oncomorphic neural agent populations for resource-limited sequential learning

Philip Greulich[1*], Michael Levin[2], Rosalia Moreddu[3*]

[1]School of Mathematical Sciences, University of Southampton, Southampton, UK
[2]Department of Biology, Tufts University, Medford, MA, USA
[3]School of Electronics and Computer Science, University of Southampton, Southampton, UK

Correspondence: p.s.greulich@soton.ac.uk, rosalia.moreddu@eng.ox.ac.uk

## Abstract

Distributed artificial intelligence (AI) often operates under sequential task exposure, uneven compute, and decentralized coordination. Here, we present a cancer-inspired, or oncomorphic, multi-agent framework in which simulated neural agents can replicate, mutate their neural network architecture, migrate across task environments, undergo ecological turnover, and recruit learning/ecological resources from a finite shared reserve. We evaluate the framework in controlled synthetic nonlinear classification environments in which each agent trains only on its local task, allowing population ecology rather than centralized optimization to determine which neural network architectures persist. For various initial conditions, we find that stronger selection increased the endpoint local accuracy of surviving agent populations. Architecture mutation played a state-dependent role: diverse initial populations performed best at low mutation, whereas clonal large-architecture populations benefited from mutation-generated variation. Selection also increased end-of-run multi-task competence, measured by evaluating surviving agents on all environments without additional training. Recruitment and elevated baseline replication reshaped demographic support while prediction quality remained within a narrow band, consistent with redistribution of finite learning resources. Time-resolved entropy and dominance analyses revealed concentration toward successful architectures, while finite training cycles kept agents in a non-asymptotic learning regime. These results provide proof-of-concept mechanistic evidence that oncomorphic population dynamics may offer a route to decentralized adaptation in engineering applications under bounded local resources.

**Keywords:** oncomorphic computing; multi-agent systems; neural agents; sequential learning; evolutionary computation; decentralized adaptation.

## 1. Introduction

The biological world has long been a source of inspiration for computing systems[1, 2]. Swarm intelligence demonstrated that local rules generate collective problem solving, as in particle swarm and ant colony optimization[3], as well as slime-mold problem-solving[4]. Evolutionary computation established that variation, inheritance, and differential success can search complex spaces effectively[5]. Artificial immune systems contributed powerful ideas about diversity maintenance, distributed anomaly detection, and adaptive memory[6]. These findings transformed how we understand autonomy, highlighting how complex adaptive behavior can emerge from decentralized local interactions[7]. A growing body of work focuses on neural learning under sequential exposure[8, 9]. Feed-forward neural networks trained by gradient descent remain foundational in modern machine learning, yet sequential task presentation introduces a persistent challenge: learning on later tasks can perturb representations that were useful for earlier ones, producing interference[10]. Continual-learning research has employed parameter regularization, replay strategies, modular or expandable architectures, and task-sensitive inference mechanisms.[11, 12] Neuroevolution showed that neural structures can be discovered by inherited variation rather than fixed human design[13-15]. Population-based training showed that groups of models trained in parallel can be improved by periodically replacing poorly performing learners with stronger ones while adaptively tuning hyperparameters during training[16]. Neural architecture research further established that repeated evaluation and selection can identify architectures well suited to demanding tasks.[17] These approaches treat model selection as a dynamical population process[18]. Their most common form, however, still assumes a centrally managed computational budget and candidate models evaluated outside a spatially distributed ecology[7].

By contrast, a decentralized population living across multiple environments changes the conditions under which architectures persist. Edge devices, sensor networks, autonomous laboratories, cyber-physical infrastructures, and robotic collectives encounter data streams that arrive locally, tasks that change over time, and compute budgets that vary from place to place[19]. Some systems, such as cells in tissues, in effect play meta-economic games in which actions change the effective number of players and deform the payoff matrix[20]. In such systems, learning is a repeated process of local adaptation under partial information, intermittent communication, finite resources, and sequential task encounter.[21-23] This challenge raises questions on which architectures survive repeated exposure to heterogeneous tasks, how variation is introduced and filtered, how agents move across niches, and how limited support is distributed across a population of competing learners. Suitable frameworks must therefore connect four ingredients that are often studied separately: heritable variation, performance-dependent selection, mobility across environments, and finite local support[24]. In this context, cancer ecology offers an especially relevant parallel[7]. Tumor populations are heterogeneous, decentralized, strongly selective, mobile, and deeply shaped by, and shaping, their microenvironments[25-28]. They diversify through mutation, expand clonally when local conditions are favorable, disseminate across niches, and remodel support through niche construction and angiogenesis[27], demonstrating how, in nature, adaptive persistence can emerge without centralized control.

Here, we introduce the concept of oncomorphic systems to denote computational frameworks that borrow key organizational principles from cancer ecology (**Figure 1**). We model a decentralized oncomorphic population of neural agents in which learning occurs within individual agents, each carrying its own neural network and training locally on its current environment, while population-level adaptation emerges through migration, mutation, performance-dependent selection, and finite ecological carrying capacities. We then use controlled synthetic environments to characterize: (i) how selection and mutation interact under diverse versus clonal initial architecture distributions, (ii) whether local ecological selection is associated with higher end-of-run multi-task competence, and (iii) how recruitment of learning and ecological resources reshapes bounded demographic support.

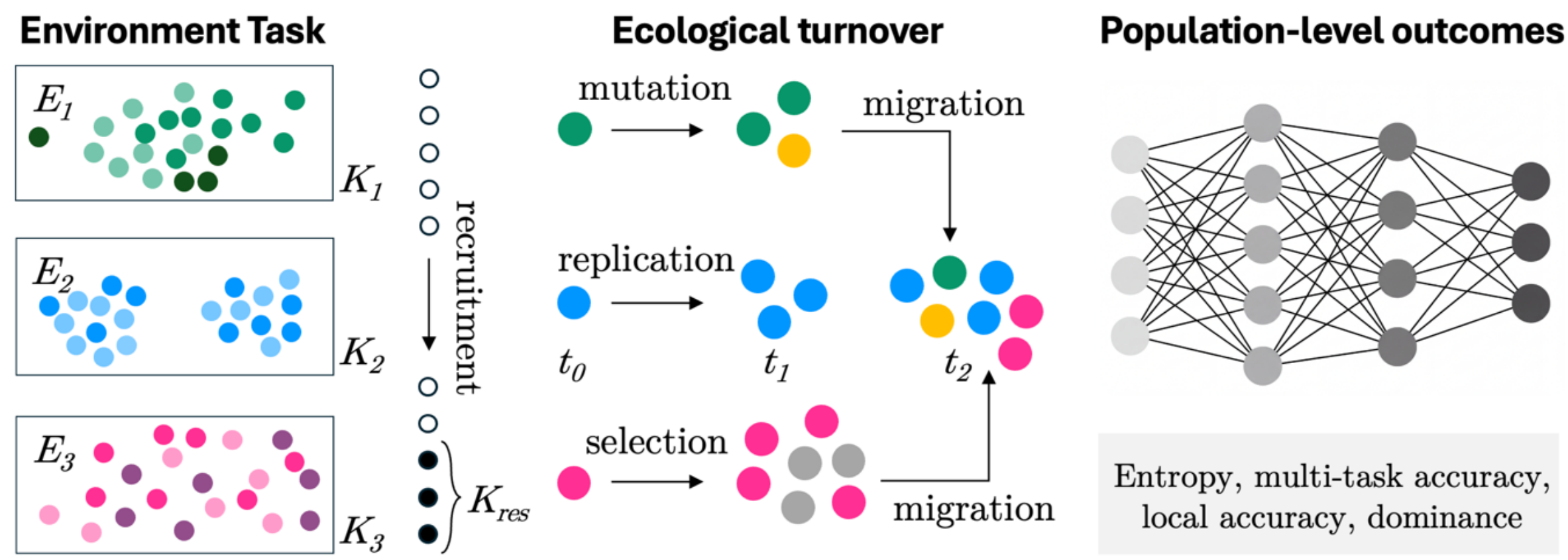

**Figure 1. Ecological turnover shapes decentralized adaptation in oncomorphic neural agent populations.** Diverse environments $E_1, E_2, E_3$ host different nonlinear classification tasks and local carrying capacities $K_1, K_2, K_3$. Neural agents, each defined by an inheritable feed-forward neural network architecture, train only on the task in the environment they currently occupy. After local learning, agents undergo ecological turnover through replication, mutation, migration, and performance-dependent selection, while migration can trigger recruitment of carrying capacity from a finite shared reserve $K_{res}$ and, after reserve depletion, by redistribution across environments. Through repeated local learning and ecological turnover, the population becomes enriched for architecture lineages that perform well under finite training budgets, producing emergent population-level outcomes including local and multi-task accuracy, dominance, and changes in architectural entropy. Colors are used to distinguish agents and lineages visually.

## 2. Results

We studied a population of individual neural agents distributed across multiple virtual environments. Each agent is a single learner: it carries one dense feed-forward neural network and trains that network only on the task of the environment it currently occupies. Each environment defines a distinct nonlinear classification task and a local carrying capacity that limits how many agents it can support. The population is the collection of all agents across environments, and population composition changes over time through replication, mutation of agent architecture, migration between environments, recruitment of carrying capacity, and performance-dependent removal. Thus, adaptation occurs on two coupled scales: within-agent learning by gradient descent during a local training episode, and between-agent ecological turnover across the population after training. Architectures vary in depth, width, activation function, and optimizer, so selection acts on locally trained agents carrying different neural-network designs rather than on centrally curated candidates. At the end of a simulation, we also evaluated each surviving trained agent on all environments without additional training to measure the final multi-task competence of the selected population. Details of the model, synthetic tasks, and simulation protocol are given in Methods.

## 2.1 Performance-based selection improves local learning, whereas the effect of mutation depends on initial architectural diversity

We first examined the joint effects of selection strength and mutation rate (**Figure 2**). Results are shown on the average local accuracy of surviving agents, for a range of mutation rates and selection pressure parameters (see Methods), for two initial conditions: a diverse random initialization (*initRand*) and a homogeneous maximal initialization (*initMax*). This comparison isolates whether mutation mainly disrupts already-available variation or generates the variation required for selection to act. Across both initial conditions, stronger performance-dependent selection was associated with higher terminal local accuracy.

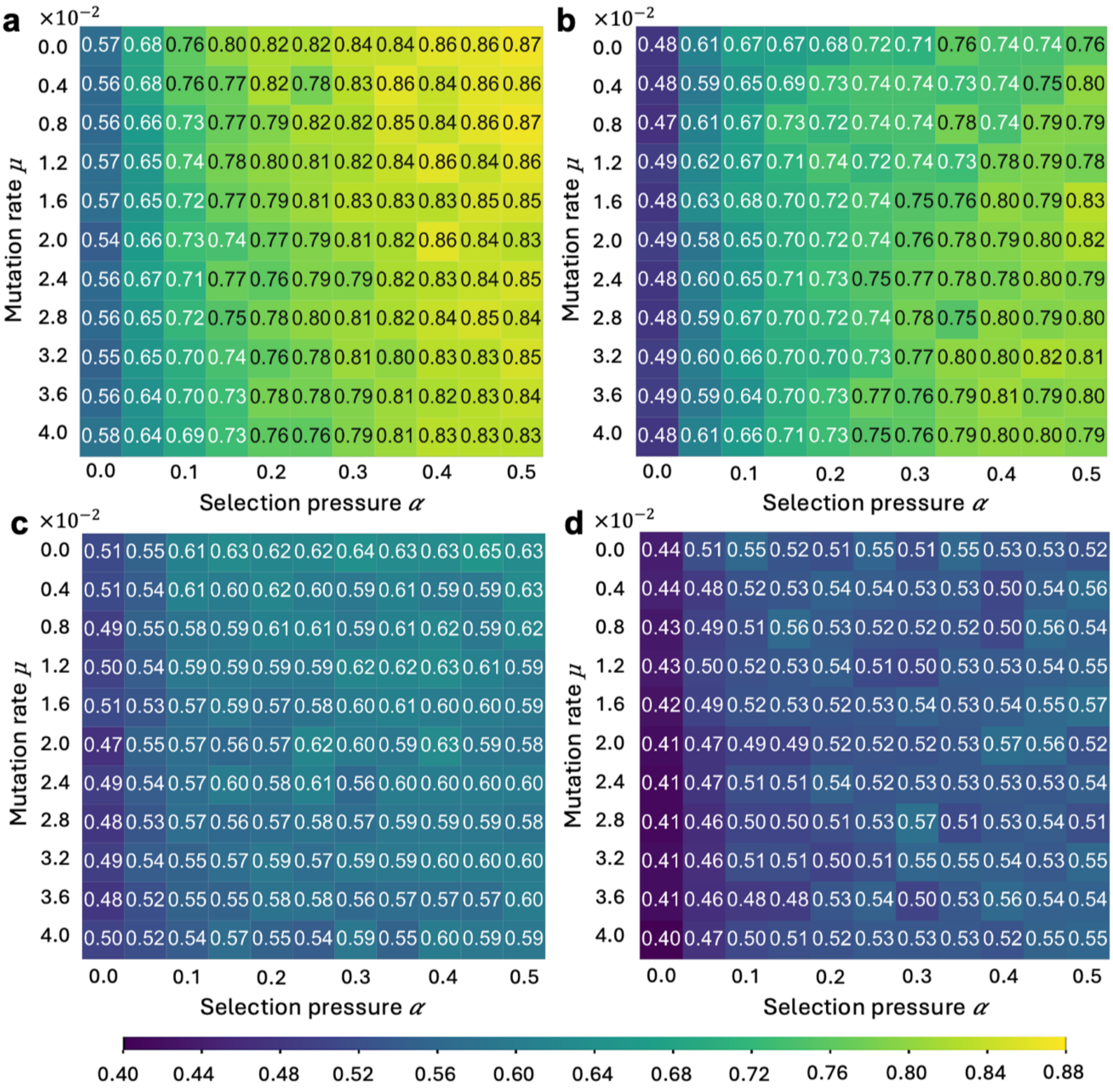

**Figure 2. Selection-mutation landscapes of local and multi-task competence performance in oncomorphic neural agent populations.** Heatmaps show mean accuracy across the sweep of selection strength α and mutation rate μ, averaged over five independent repeats. a) Single-task accuracy, *initRand*. Mean terminal local accuracy for the diverse random initial condition. Performance increases strongly with selection, with the best outcomes in the high-selection, low-mutation region, indicating that selection can effectively enrich already-available high-performing architectures. b) Single-task accuracy, *initMax*. Mean terminal local accuracy for the homogeneous maximal initial condition. Selection again improves performance, but unlike

panel a, nonzero mutation is beneficial because it creates the architectural variation needed for selection to act in an initially clonal population. c) Multi-task accuracy, *initRand*. Mean end-of-run accuracy measured by evaluating surviving agents on all environments without additional training. Although absolute values are lower than in panel a, the same general trend appears: stronger selection improves performance, and lower mutation performs best when initial diversity is already present. d) Multi-task accuracy, *initMax*. Mean end-of-run multi-task accuracy for the clonal maximal initial condition. As in panel b, performance improves with selection and benefits from nonzero mutation, showing that mutation is constructive when initial architectural diversity is absent.

In *initRand*, mean local accuracy increased from approximately 0.577 when mutation and selection are absent to approximately 0.878 at high selection with zero mutation (**Figure 2a,b**). Much of the high-selection region exceeded 0.84. This pattern indicates that, when standing architectural variation is already present, selection enriches architectures that learn their local task more effectively under the available epoch budget. In this diverse initial condition, mutation had a limited and often negative marginal effect once selection was active. At selection strength 0.5, mean local accuracy was approximately 0.878 at zero mutation and decreased to about 0.833 at the highest mutation level shown. Thus, in the presence of substantial standing variation, continued perturbation appears to oppose consolidation around well-performing architectures. The behavior differed in *initMax*, in which all agents began from the same large architecture (8 hidden layers, 256 nodes per layer). Here, selection alone improved performance, but mutation became constructive because it generated the variation on which selection could act. Mean local accuracy increased from approximately 0.480 with no selection and no mutation to roughly 0.765 with strong selection and zero mutation and rose further to about 0.834 at nonzero mutation in the high-selection region. In this clonal setting, mutation supplies architectural alternatives, and selection amplifies those with better finite-budget learning behavior. These results indicate that selection is the main driver of improved local performance, whereas the effect of mutation is state-dependent. Mutation is beneficial when useful variation is initially absent, but can be disruptive when selection is already acting on a broad and favorable architecture distribution.

## 2.2 Selection enriches multi-task competence among surviving agents under sequential local training

We next evaluated whether local ecological selection also altered end-of-run competence across all environments (**Figure 2c,d**). For this analysis, each surviving agent was evaluated on every environment simultaneously without additional training. This metric therefore measures retained multi-task competence rather than further task-specific adaptation. In *initRand*, mean multi-task accuracy increased from approximately 0.511 at zero selection and zero mutation to values in the 0.60 range, with a peak near 0.651. Although these values were lower than the corresponding local accuracies, the direction of change was consistent: stronger selection on local performance was associated with improved end-of-run competence across the full task set. In the high-selection region, the best values again occurred at low mutation. At selection strength 0.5, mean multi-task accuracy was approximately 0.632 at zero mutation and closer to 0.594 at the highest mutation shown. In *initMax*, the baseline was lower, beginning near 0.440 at zero selection and zero mutation. As in the local-accuracy analysis, nonzero mutation improved outcomes under selection, with the best values in the high-selection region reaching approximately 0.57. Thus, when the initial population is clonal, mutation again appears to provide the architectural alternatives required for ecological enrichment. These results do not by themselves establish broad transfer in the continual-learning sense. However, they do show that a population shaped only by local training, yet sequentially on different tasks, and ecological turnover can become enriched for architectures associated with higher end-of-run multi-task competence across related synthetic environments drawn from the same task family.

## 2.3 Recruitment reshapes demographic support while prediction quality remains bounded

We then examined whether local recruitment of learning and ecological carrying capacity, and changes in baseline replication rate, altered prediction quality at the population level (**Figure 3**). In that scenario, agents could at each learning event increase the carrying capacity of their environment, $K_e$ (the number of agents the environment can maximally host), which also serves as course for learning capacity, assuming that the number of learning epochs available to each agent are proportional to $K_e$. Across recruitment probability and replication rate values, mean local accuracy remained in a relatively narrow range, approximately 0.785–0.853, and no clear monotonic trend was apparent along either parameter axis. This bounded response suggests that intensified demographic activity is not equivalent to systematically improved prediction quality. In the present model, recruitment first draws $K_e$ from an external reserve and later from the carrying capacities of other environments. Recruitment therefore acts as a redistribution mechanism rather than as unbounded resource creation. An environment that gains support can offer a higher local training budget (number of epochs), but that gain is eventually offset elsewhere once the reserve is depleted. Accordingly, selection and mutation primarily determine which architectures persist, whereas recruitment and baseline replication mainly alter how finite support is distributed across the ecology.

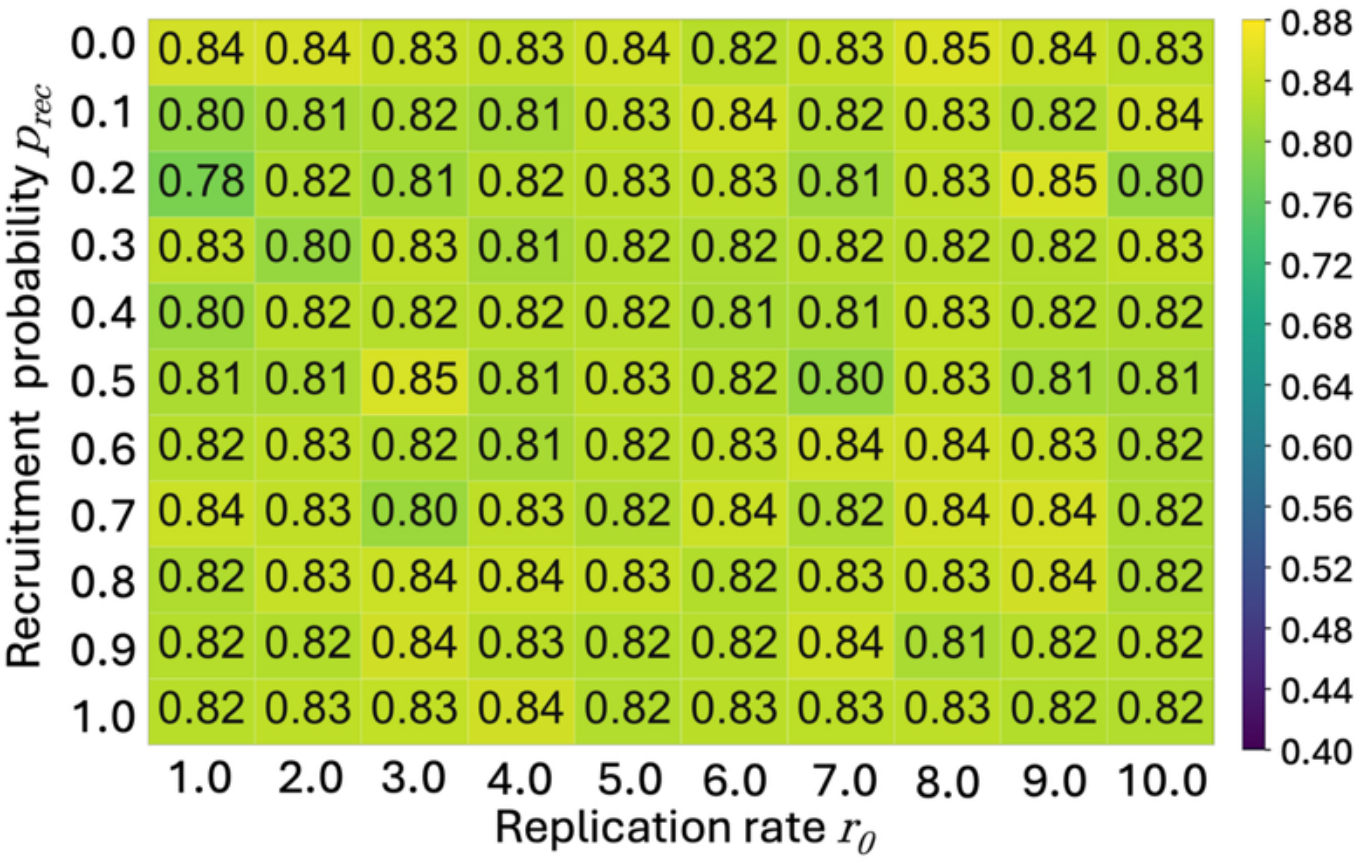

**Figure 3. Recruitment and baseline replication have limited effects on local prediction accuracy.** Heatmap of mean terminal local accuracy in the random initial condition across baseline replication rate $r_0$ and recruitment probability $p_{rec}$. Accuracy remains within a narrow range across the parameter sweep, indicating that these processes mainly redistribute finite ecological and learning support rather than substantially improving prediction performance.

## 2.4 Time-resolved dynamics reveal selection-driven concentration, diversification from clonal starts, and finite-cycle learning

Endpoint heatmaps identify favorable parameter regions, but the time-resolved analyses clarify how those outcomes arise (**Figure 4**). In the neutral *initRand* control, the dominant-architecture fraction increased only modestly over time, reaching roughly 0.08, while the Shannon entropy[29], measuring architectural diversity, declined from about 4.1 to 3.75. Thus, some concentration emerges even in the absence of selection, consistent with stochastic drift in a finite population[30]. Under selection and mutation in *initRand* (**Figure 4a**; α = 0.4, μ = 0.02, chosen as parameter values for which accuracy was maximal), concentration was markedly stronger. The dominant-architecture fraction increased to approximately 0.20–0.23 and entropy decreased by about 1.2 over the run. At the repeat-averaged level, the most frequent architecture was a relatively compact network with only two hidden layers, 140 nodes per layer, ReLU activation, and RMSprop optimization. Representative runs showed stronger local dominance than the repeat average, indicating that selection reproducibly concentrates the population toward a restricted region of architecture space even if exact rank order varies between realizations. Notably, the selected architectures are not those with the maximal depth or width among those present initially.

The complementary pattern appeared in *initMax* (**Figure 4b**). Because the initial population was clonal, the dominant-architecture fraction started at 1 and entropy at 0. With mutation active, dominance declined, and entropy increased to approximately 2.4, indicating diversification via mutation away from the founding architecture. Selection then filtered this newly generated variation, yielding a distributed set of lineages that outperformed the original clonal state. In a diverse population, mutation competes with consolidation by perturbing already favorable designs. In a clonal population, mutation is the route by which selection gains access to alternatives. The mutation process is therefore exploratory in one regime and disruptive in another. **Figure 4c** provides a final within-cycle view by examining learning dynamics within individual training cycles. The gap between the loss at epoch τ and the final within-cycle loss decayed steadily toward zero. This indicates that agents continue to improve throughout a local episode rather than reaching immediate saturation. It also suggests that, under the default settings, agents are typically evaluated, selected, and redistributed before exhaustive convergence on the current task. Under these conditions, architecture quality depends not only on asymptotic capacity but also on trainability under short or moderate learning windows. The temporal analyses support the interpretation of the heatmaps as an ecological process of architecture selection. Neutral drift alone produces only limited concentration, whereas performance-dependent selection generates stronger and more reproducible contraction toward a subset of architecture space.

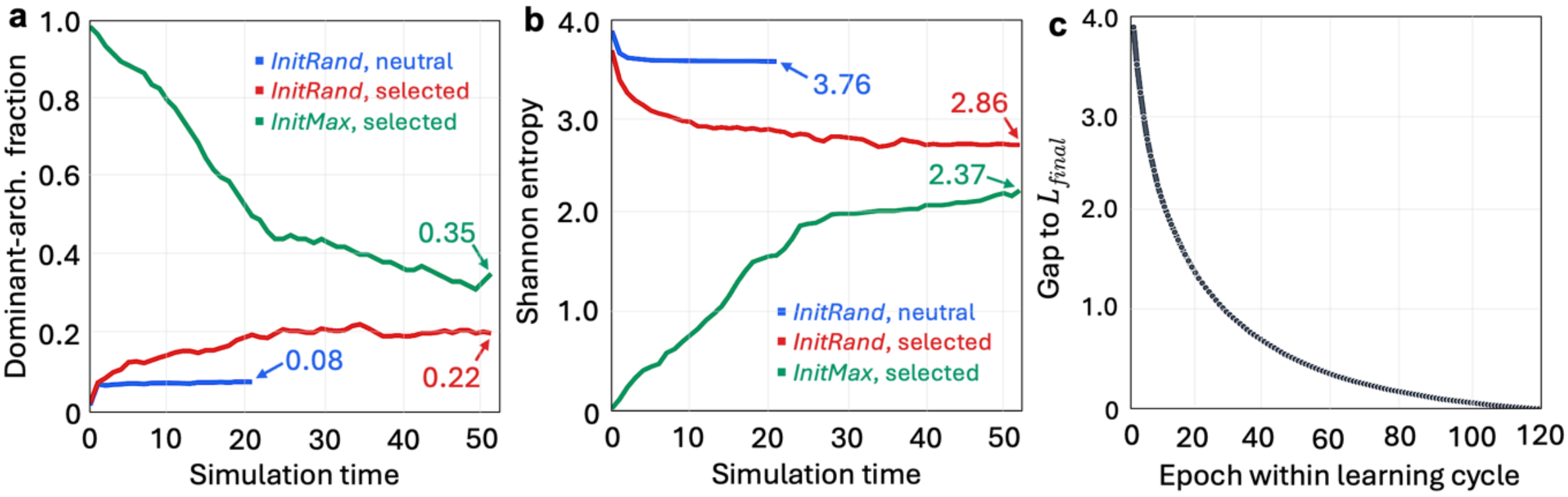

**Figure 4. Time-resolved signatures of ecological selection, diversification from clonal starts, and finite-cycle learning in oncomorphic neural agent populations. a)** Dominant-architecture fraction as a function of simulation time for the three regimes used in the temporal comparison: neutral (μ=0, α=0) *initRand* (blue), selected (μ=0.02, α=0.4) *initRand* (red), and

selected *initMax* (green). This metric reports the population share occupied by the most frequent architecture and therefore quantifies concentration of the architecture distribution over time. **b)** Shannon entropy of the architecture distribution for the same three regimes, providing the complementary measure of architectural diversity. Together, panels (a) and (b) show weak drift-driven concentration in the neutral control, stronger contraction under selection from a diverse starting population, and mutation-enabled diversification from the *initMax* condition. **c)** Within-cycle learning trajectory, plotted as the gap between the loss at epoch τ and the final within-cycle loss versus epoch. The progressive decay of this gap indicates that agents continue to improve during local training episodes and are evaluated, selected, and redistributed before full convergence is reached.

## 2.5 Endpoint abundance structure of the most frequent architectures reveals how selection reshapes the occupied architecture distribution

To examine the endpoint population structure more directly, we analyzed the endpoint profiles of the eight most frequent architectures in the same three regimes used for the temporal comparisons (**Figure 5**). **Figure 5a** shows the repeat-averaged endpoint distribution, highlighting concentration patterns that recur across independent realizations. **Figure 5b** shows a representative single-run endpoint distribution, which resolves the stronger top-heaviness that can emerge within individual stochastic realizations. These results distinguish reproducible redistribution of population mass from run-to-run variation in exact architecture identity. In the neutral *initRand* control, abundance remained comparatively spread across the displayed architectures, consistent with the modest increase in dominant-architecture fraction and the limited entropy decline observed earlier in *Figure 4*.

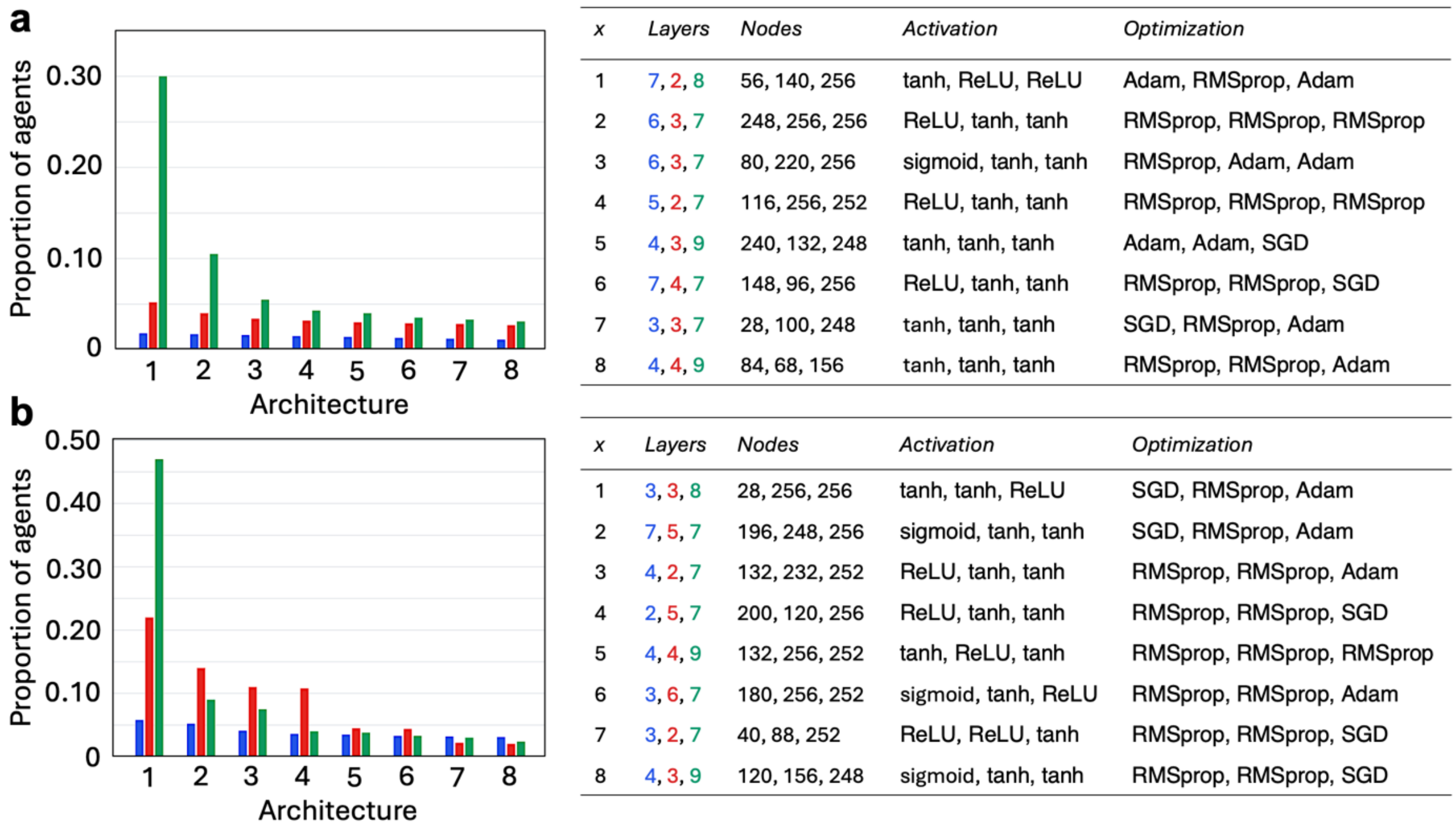

| *x* | *Layers* | *Nodes* | *Activation* | *Optimization* |
|---|---|---|---|---|
| 1 | 7, 2, 8 | 56, 140, 256 | tanh, ReLU, ReLU | Adam, RMSprop, Adam |
| 2 | 6, 3, 7 | 248, 256, 256 | ReLU, tanh, tanh | RMSprop, RMSprop, RMSprop |
| 3 | 6, 3, 7 | 80, 220, 256 | sigmoid, tanh, tanh | RMSprop, Adam, Adam |
| 4 | 5, 2, 7 | 116, 256, 252 | ReLU, tanh, tanh | RMSprop, RMSprop, RMSprop |
| 5 | 4, 3, 9 | 240, 132, 248 | tanh, tanh, tanh | Adam, Adam, SGD |
| 6 | 7, 4, 7 | 148, 96, 256 | ReLU, tanh, tanh | RMSprop, RMSprop, SGD |
| 7 | 3, 3, 7 | 28, 100, 248 | tanh, tanh, tanh | SGD, RMSprop, Adam |
| 8 | 4, 4, 9 | 84, 68, 156 | tanh, tanh, tanh | RMSprop, RMSprop, Adam |

| *x* | *Layers* | *Nodes* | *Activation* | *Optimization* |
|---|---|---|---|---|
| 1 | 3, 3, 8 | 28, 256, 256 | tanh, tanh, ReLU | SGD, RMSprop, Adam |
| 2 | 7, 5, 7 | 196, 248, 256 | sigmoid, tanh, tanh | SGD, RMSprop, Adam |
| 3 | 4, 2, 7 | 132, 232, 252 | ReLU, tanh, tanh | RMSprop, RMSprop, Adam |
| 4 | 2, 5, 7 | 200, 120, 256 | ReLU, tanh, tanh | RMSprop, RMSprop, SGD |
| 5 | 4, 4, 9 | 132, 256, 252 | tanh, ReLU, tanh | RMSprop, RMSprop, RMSprop |
| 6 | 3, 6, 7 | 180, 256, 252 | sigmoid, tanh, ReLU | RMSprop, RMSprop, Adam |
| 7 | 3, 2, 7 | 40, 88, 252 | ReLU, ReLU, tanh | RMSprop, RMSprop, SGD |
| 8 | 4, 3, 9 | 120, 156, 248 | sigmoid, tanh, tanh | RMSprop, RMSprop, SGD |

**Figure 5. Endpoint profiles of top eight architectures across control and selected regimes.** **a)** Repeat-averaged final proportions of agents occupying the top eight architectures in the neutral *initRand* control (blue; α=0, μ=0, $p_{rec}$=0), selected *initRand* (red; α=0.4, μ=0.02, $p_{rec}$=0), and selected *initMax* (green; α=0.4, μ=0.02, $p_{rec}$=0). The progressively steeper profiles from blue to red to green indicate increasingly strong concentration of population mass into a small

number of architectures. The table at the right of each plot lists the corresponding architecture specifics in the three regimes in terms of layer number, nodes per layer, activation function, and optimizer. Color order is indicated in the layers column. **b)** Representative single-run endpoint distributions for the same three regimes, together with the corresponding architecture specifics. Relative to the repeat-averaged view in panel a, individual realizations show stronger top-heaviness while preserving the same qualitative ordering across regimes. In both panels, x=1–8 denotes the display index corresponding to the table rows.

Thus, in the absence of performance-dependent selection, finite-population drift produces only weak compression of the occupied architecture distribution. On the contrary, the selected *initRand* regime exhibited a more concentrated abundance profile, indicating that ecological filtering redistributes population mass toward a restricted set of recurrently successful architectures. This endpoint structure is the distributional counterpart of the stronger concentration and entropy reduction seen for the red regime in *Figure 4*. The selected *initMax* regime displayed the strongest twist: starting from a clonal large architecture, mutation first generates alternatives, and selection subsequently filters those alternatives into a smaller successful subset, yielding a strongly top-heavy endpoint distribution. This pattern matches the green trajectories in *Figure 4*, where diversity is created from an initially clonal state and then restricted by ecological selection. The tables identify the layer number, node number, activation function, and optimizer associated with each displayed architecture in the blue, red, and green regimes.

## 3. Discussion

In the presented model, AI agents represented by feed-forward neural networks undergo architectural mutation, clonal expansion, migration across environments (tasks), performance-conditioned selection, and local recruitment of finite learning and ecological resources. Within this setting, ecological turnover enriches architecture classes that learn effectively under heterogeneous and bounded training conditions. The central result is that stronger ecological selection was associated with higher local accuracy and better final performance across the full set of environments, even though each agent trained only on the task in its current environment. In other words, populations exposed only to local sequential learning became enriched for agent architectures that, by the end of the simulation, performed better not just on their current task but also when tested across all tasks without additional training. This finding suggests that, in a distributed multi-agent setting, useful competence need not arise solely from preserving all knowledge within a single persistent learner. It may also arise from population-level enrichment of architectures that repeatedly perform well under sequential local exposure. The second outcome is that the effect of mutation depends strongly on the initial geometry of the architecture distribution. When the population begins with broad standing variation, low mutation favors consolidation around successful designs. When the population begins from a clonal architecture, mutation becomes necessary to generate alternatives and thereby enable selection. This state dependence is conceptually important because it suggests that exploratory pressure in decentralized AI systems may be most useful when diversity is low, and less useful once the ecology has already explored favorable regions of design space. Relative to standard architecture search or population-based optimization, the distinguishing feature of the present framework is that candidate architectures persist or disappear through repeated local learning under migration, bounded carrying capacity, and ecological turnover.

The multi-task competence results warrant further investigation. The present experiments show that local ecological selection can enrich architectures associated with better end-of-run performance across related environments without additional training. However, future work should incorporate standard continual-learning metrics such as backward transfer, forward transfer, task-wise forgetting, and calibration under migration for a more definitive assessment. The recruitment analysis further indicates that local amplification of support is not equivalent to population-wide performance gains when total resources are bounded. In the current model, recruitment is eventually redistributive: increasing carrying capacity in one environment reduces available support elsewhere. This distinction between adaptation and allocation is critical. Selection and mutation mainly influence which architectures persist, whereas recruitment and baseline replication reshape the demographic substrate on which learning takes place. The time-resolved analyses also help distinguish ecological selection from stochastic drift. Drift alone produced only modest changes in dominance and entropy, whereas selection led to stronger and more reproducible concentration. Notably, the architectures that rose to high frequency were often relatively compact rather than maximally large. Under finite local training budgets, trainability and efficiency can outweigh raw capacity. This observation is relevant to edge and embodied settings, where time, memory, and energy constraints may favor architectures that learn quickly over those that are only superior in the long training limit[31].

The main limitations of the study are the use of synthetic task families and architecture-level inheritance. Accordingly, the results should be interpreted as mechanistic evidence that ecological turnover can shape architecture distributions under bounded local training. The model opens a number of future directions. The same logic could be tested with convolutional, recurrent, graph-based, or transformer-like agents, with explicit cost models for latency, parameter count, energy, and communication, and with richer data modalities including images, time series, language, or multimodal streams. Weight inheritance, partial parameter transfer, and recombination would also allow direct study of how architectural and parametric memory interact under ecological turnover. Relative to most existing continual-learning, neuroevolution, neural architecture search, and population-based training studies, this framework shifts the focus from centralized optimization to decentralized ecological turnover, in which local learning, migration, mutation, and ecological selection under bounded resources jointly enrich successful architectures at the population level.

# 4. Methods

## 4.1 Oncomorphic agent representation and local learning

The model comprises a finite population of neural-network agents distributed across a set of classification environments $E = \{1, \dots, n_{env}\}$. At time $t$, agent $i$ occupies environment $e_i(t) \in E$ and carries an inheritable architecture

$$A_i = (m_i, h_i, \phi_i, o_i)$$

where $m_i$ is the number of hidden layers, $h_i$ is the number of nodes per hidden layer, $\phi_i$ is the hidden-layer activation function, and $o_i$ is the optimizer. Each architecture defines a dense feed-forward network

$$f_{\theta_i, A_i}: \mathbb{R}^{d_x} \to \mathbb{R}^{d_y}$$

with trainable parameters $\theta_i$. The study focuses on dense feed-forward models because they provide a transparent architecture space in which local changes to depth, width, activation, and optimizer can be interpreted directly as heritable mutations. Allowed activation functions were ReLU, tanh, and sigmoid. Allowed optimizers were Adam, RMSprop, and stochastic gradient descent. The heritable object in the population is therefore a compact architecture-optimizer specification instead of a monolithic fixed model class.

Learning occurs locally. Each environment supplies one classification task, and an agent trains only on the task of its current environment. Single-label tasks use softmax outputs with categorical cross-entropy loss. The orthant task uses sigmoid outputs with binary cross-entropy. In all cases, learning within a lifetime follows standard gradient-based supervised optimization in the sense of feed-forward neural learning[9]. The defining ecological feature is that the duration of training is environment dependent. Each environment $e$ has a carrying capacity $K_e(t)$ and the number of epochs available during a learning episode is

$$T_e(t) = \eta K_e(t)$$

rounded to an integer, where $\eta$ denotes epochs per unit carrying capacity. Carrying capacity therefore acts as a local computational support variable: environments with larger support offer more learning time, while more constrained environments enforce shorter training cycles. This coupling between ecology and optimization is central to the framework because it ensures that architecture quality is assessed under finite, location-specific learning budgets rather than under unlimited convergence.

Architectural inheritance is tracked at the population level. Replication creates an offspring carrying the parent architecture, allowing successful design classes to expand through the ecology. Mutation replaces an agent with a newly initialized neighboring architecture, making architecture rather than precise parameter state the principal heritable substrate. This separation isolates selection over architecture classes and allows the results to be interpreted as architecture-level meta-selection rather than as ordinary weight inheritance. In the simulations reported here, inheritance operated at the architecture level only: offspring and mutants were newly initialized and did not inherit parental weights. The population is therefore composed of locally learning agents whose lifetime adaptation is handled by gradient descent, while cross-lifetime adaptation is handled by ecological turnover over architecture space.

## 4.2 Virtual nonlinear environments, label generation, and evaluation of accuracy

All tasks were generated as controlled virtual classification problems, with training and test sets created independently for each environment. Inputs were sampled from standard normal distributions in $\mathbb{R}^{d_x}$. For each environment e and output component $k \in \{1, \dots, d_y\}$ , we chose a latent nonlinear response function that is bounded and allows for periodicity. We chose to define this function as

$$g_{e,k}(x) = \left[ tanh\left(\sum\nolimits_{l} \omega_{e,k,l} x_l\right) + a\ sin\left(\sum\nolimits_{l} \tilde{\omega}_{e,k,l} x_l\right) \right] + b_{e,k} + \chi$$

Where $\omega_{e,k,\ell}$, $\tilde{\omega}_{e,k,\ell}$ are linear coefficients, $a$ is the weighting ratio, $b_{e,k}$ an offset term and $\chi$ a stochastic noise term. The coefficients and bias were sampled independently from a standard normal distribution (with zero mean and unit variance) for each environment and output component, producing environment-specific nonlinear structure. The noise term $\chi$ was sampled for each instance of $x$, with mean zero and standard deviation 0.1. The value of $a$ was kept fixed at $a=0.4$. The latent vector for environment $e$ was thus,

$$z_e(x) = \left(g_{e,1}(x), \dots, g_{e,d_y}(x)\right)$$

Labels were derived from these latent variables using three task families. In the argmax task, the target class is the index of the largest latent component,

$$y(x) = \arg\max_{k \in \{1,\dots,d_y\}} z_{e,k}(x)$$

This produces a nonlinear multi-class problem defined by comparisons among latent responses. In the Gaussian-blob task, randomly generated class centers in latent space define noisy clusters, with center coordinates sampled from normal distributions and variance $\sigma^2 = 4$. This retains single-label classification while emphasizing cluster structure after the nonlinear transformation. In the orthant task, each output dimension defines a binary label according to the sign of the corresponding latent response,

$$y_k(x) = 1\{z_{e,k}(x) > 0\}, \qquad k = 1, \dots, d_y$$

This creates a multi-label problem requiring simultaneous learning of several nonlinear sign boundaries.

Performance was measured after each local training episode. For single-label tasks, the agent-level accuracy $q_i \in [0, 1]$ is the fraction of correctly classified test samples in the agent current environment. For the orthant task, $q_i$ is the mean component-wise binary accuracy across examples and output dimensions. The associated classification error is

$$\varepsilon_i = 1 - q_i$$

This local accuracy enters the ecological dynamics through a sigmoidal performance component,

$$s_i = \frac{1}{1 + \exp\left(\alpha \varepsilon_i\right)}$$

where $\alpha$ is the selection-strength parameter. Larger $\alpha$ values increase ecological sensitivity to performance differences. In effect, local prediction quality is transformed into a selective advantage that acts jointly with carrying-capacity constraints. We used this form because it provides a smooth, bounded, monotone transformation of classification error into ecological advantage, preventing large rate excursions while allowing α to tune the sharpness of performance-based discrimination.

Two levels of predictive outcome were analyzed. The first is average single-task accuracy, obtained by averaging the most recent local test accuracies of surviving agents. The second is average multi-task accuracy, denoted $q_i^{MT}$, computed at the end of each simulation by evaluating each trained agent on all environments without additional training. This second measure is essential because it distinguishes narrow local adaptation from broader retained competence across the task family.

## 4.3 Ecological event dynamics, carrying-capacity recruitment, and stochastic simulation

Agents participate in a continuous-time ecological process comprising replication, mutation, migration, recruitment, and performance-conditioned removal. Let $N_e(t)$ denote the number of agents currently occupying environment $e$ and its carrying capacity, $K_e(t)$. The replication rate of agent $i$ follows a logistic law:

$$\lambda_i^{rep}(t) = r_0 \left(1 - \frac{N_e(t)}{K_e(t)}\right) s_i(t)$$

where $r_0$ is a baseline replication parameter and $s_i(t)$ is the performance-dependent term defined from the most recent local accuracy. Replication therefore has both ecological and performance components: it is facilitated by under-occupied environments and by stronger learned performance. In implementation, replication rates were clipped at zero whenever $N_e(t) > K_e(t)$, ensuring that all Gillespie event rates remained non-negative.

Mutation acts locally in architecture space. If agent $i$ carries architecture $A_i = (m_i, h_i, \phi_i, o_i)$, mutation generates a neighboring architecture by perturbing depth and width and resampling categorical hyperparameters. The layer count is updated as:

$$m_i' = max\{1, m_i + \Delta m\}, \qquad \Delta m \in \{-1, +1\}$$

and width is updated as

$$h_i' = clip(h_i + \Delta h, 8{,}256), \quad \Delta h \in \{-8, -4, +4, +8\}$$

Activation function and optimizer are resampled from the allowed sets. This mutation operator keeps exploration local and interpretable. Architectures evolve through adjacent changes rather than through arbitrary global resampling, which makes selected trajectories meaningful in terms of gradual ecological search.

Migration moves an agent to a randomly chosen environment at rate ν, and we also chose this rate to fix the simulation time unit. Recruitment occurs probabilistically after migration with probability $p_{rec}$. A recruitment event increases the carrying capacity of the receiving environment by $\Delta K$. Capacity is drawn first from an external reserve $K_{res}$. After reserve depletion, additional recruitment transfers capacity from other environments, so local gains are balanced by reductions elsewhere.

Selection is implemented as performance-dependent removal after training. An agent with local accuracy $q_i$ is removed with probability

$$p_i^{sel} = 1 - q_i^{\alpha}$$

Larger $\alpha$ values therefore increase the likelihood that weakly performing agents are removed after a learning episode, sharpening ecological filtering.

All events were simulated as a continuous-time Markov chain using the Gillespie algorithm[32]. This was chosen because the model is defined by event rates rather than synchronized discrete generations. The Gillespie procedure samples both the time to the next event and the event identity from the current rate structure, preserving the stochastic ordering of ecological dynamics[32]. As a result, learning, turnover, movement, and support reallocation unfold asynchronously. The output therefore captures both endpoint statistics and the temporal organization of adaptation.

## 4.4 Experimental design, initial conditions, parameter sweeps, and quantitative readouts

Two initial conditions were used to separate the role of standing variation from mutation-generated variation. In the random initialization (*initRand*), agent architectures were sampled uniformly from the mutation-accessible space, that is, each chosen with equal probability. This created an architecturally diverse starting population. In the homogeneous maximal initialization (*initMax*), all agents began with the same architecture containing the maximal values of eight hidden layers and 256 nodes per layer. This condition provided a clonal starting state with maximal initial width and high depth, allowing direct evaluation of whether mutation is necessary to create useful alternatives when architecture diversity is absent. Unless stated otherwise, simulations used the default parameter values listed in **Table 1**. The migration rate ν defined the time unit as 1/ν, so all other rates are interpreted relative to migration. Selection-mutation parameter sweeps varied selection strength α from 0 to 0.5 and mutation rate μ from 0 to 0.04. Each grid point was averaged over five independent repeats, and the resulting means are shown in *Figure 2* and *Figure 3*. Recruitment-replication parameter sweeps varied baseline replication rate $r_0$ from 1 to 10 and recruitment probability $p_{rec}$ from 0 to 1, using the same multi-repeat protocol. Time-resolved analyses focused on the neutral control ($\alpha = 0, \mu = 0, p_{rec} = 0$), and a high-performing selected regime ($\alpha = 0.4, \mu = 0.02, p_{rec} = 0$)..

**Table 1.** Simulation parameters.

| *Parameter* | *Value* | *Interpretation* |
|---|---|---|
| $p_{rec}$ | 0 | Recruitment probability |
| $\Delta K$ | 5 | Carrying-capacity increment per recruitment event |
| $K_{res}$ | 50 | External reserve capacity |
| $K_e(0)$ | 30 | Initial carrying capacity per environment |
| $\eta$ | 0.5 | Epochs per unit carrying capacity |
| ν | 1.0 | Migration rate / time-scale reference |
| $t_{end}$ | 50 | Simulation end time |
| $n_{env}$ | 6 | Total number of environments (2 per each task type) |
| $N_e(t = 0)$ | 10 | Initial number of agents per environment |
| $d_x$ | 10 | Length of virtual data input vector *x* |
| $d_y$ | 5 | Length of virtual data output vector *y* |
| $n_{terms}$ | 10 | Number of non-linear terms for generating *y* |
| $N_{train}$ | 200 | Number of training data points |

| $N_{test}$ | 100 | Number of validation data points |
|---|---|---|

The principal outcome measures were defined at the population level. Average single-task accuracy is the mean of the most recent local accuracies $q_i$ of surviving agents. Average multi-task accuracy is the mean of $q_i^{MT}$, computed by inference across all environments without additional training. Architecture composition was summarized by the frequency distribution $p_A$ over architectures, the dominant-architecture proportion $p_d = \max_A p_A$, and the Shannon entropy, defined as

$$H(t) = -\sum_A p_A(t) \log p_A(t)$$

which quantifies architecture diversity over time[28]. Time-resolved plots tracked $p_d(t)$, $H(t)$, and top-ranked architecture frequencies. For the architecture distributions, both repeat-averaged distributions and representative single-run distributions were examined. This combination is useful because repeat averages reveal reproducible concentration at the feature level, whereas single runs show the extent to which exact rank order varies across stochastic realizations. The overall design was therefore built to answer four linked questions: whether selection improves local learning, whether this effect depends on initial diversity, whether local selection improves multi-task competence, and how resource recruitment shapes ecological support.

## 5. Conclusion

In this work, neural agents were endowed with cancer-inspired properties (architectural mutation, clonal expansion, migration across environments, performance-conditioned selection, and local recruitment of resources). These properties were embedded within a resource-limited sequential learning model. Results produced a preliminary pattern of adaptation. Selection strengthened local task performance, enriched broader multi-task competence without explicit joint optimization, and concentrated the population into recurrently successful neural network architectures. Mutation contributed most strongly when initial architectural diversity was absent, while diverse starting populations benefited from lower mutation and stronger consolidation. Recruitment and baseline replication reshaped demographic support, but did not improve overall performance, clarifying the distinction between local advantage and population-level performance. Time-resolved analyses showed that the selected architectures were often compact and efficient rather than maximal in scale, and that finite learning cycles placed the system in a regime where trainability and retained competence mattered jointly. These results support oncomorphic computing as a potential proof-of-concept design space for decentralized AI under bounded local resources.

## Data availability statement

The codes are available on GitHub: https://github.com/philipgreulich/AI-agents. The data is available from the corresponding authors upon request.